\documentclass[10pt,twocolumn,letterpaper]{article} 

\usepackage{avss}
\usepackage{times}
\usepackage{epsfig}
\usepackage{graphicx}
\usepackage{amsmath}
\usepackage{amssymb}
\usepackage{algorithm2e}
\usepackage{float}
\usepackage{subfigure}
\usepackage{placeins}

\avssfinalcopy 

\def\avssPaperID{75}

\ifavssfinal\fi
\begin{document}

\title{On the Interaction Between Deep Detectors and Siamese Trackers \\in Video Surveillance}

 
\author{Madhu Kiran$^a$, Vivek Tiwari$^{b}$, Le Thanh Nguyen-Meidine$^a$, Louis-Antoine Blais Morin$^{c}$, Eric Granger$^a$\\
$^a$ \small{\textit{Laboratoire d'imagerie, de vision et d'intelligence artificielle, \'Ecole de technologie sup\'erieure, Montreal, Canada}}\\
$^b$ \small{\textit{Indian Institute of Technology(ISM), Dhanbad, India}}\\
$^c$ \small{\textit{Genetec Inc., Montreal,Canada}}\\
{\tt\small mkiran@livia.etsmtl.ca,tiwarivivekraj17@gmail.com,lethanh@livia.etsmtl.ca} \\
{\tt\small lablaismorn@genetec.com, eric.granger@etsmtl.ca}
}

\maketitle
\thispagestyle{empty}

\begin{abstract}
\vspace*{-0.5cm}

Visual object tracking is an important function in many real-time video surveillance applications, such as localization and spatio-temporal recognition of persons. In real-world applications, an object detector and tracker must interact on a periodic basis to discover new objects, and thereby to initiate tracks. Periodic interactions with the detector can also allow the tracker to validate and/or update its object template with new bounding boxes. However, bounding boxes provided by a state-of-the-art detector are noisy, due to changes in appearance, background and occlusion, which can cause the tracker to drift. Moreover, CNN-based detectors can provide a high level of accuracy at the expense of computational complexity, so interactions should be minimized for real-time applications. 
In this paper, a new approach is proposed to manage detector-tracker interactions for trackers from the Siamese-FC family. By integrating a change detection mechanism into a deep Siamese-FC tracker, its template can be adapted in response to changes in a target's appearance that lead to drifts during tracking. An abrupt change detection triggers an update of tracker template using the bounding box produced by the detector, while in the case of a gradual change, the detector is used to update an evolving set of templates for robust matching. 
Experiments were performed using state-of-the-art Siamese-FC trackers and the YOLOv3 detector on a subset of videos from the OTB-100 dataset that mimic video surveillance scenarios. Results highlight the importance for reliable VOT of using accurate detectors. They also indicate that our adaptive Siamese trackers are robust to noisy object detections, and can significantly improve the performance of Siamese-FC tracking. 
\end{abstract}

\section{Introduction}

Visual Object Tracking (VOT) is an important component in many video surveillance applications to localize objects and persons, and possibly regroup their images, for further processing in applications such as scene understanding, action recognition, person re-identification, expression recognition~\cite{Miguel, Dewan}. Some of the challenges faced by VOT in such real-world applications are changes in pose, illumination, occlusion, deformation, motion blur and complex backgrounds. Additionally, in real-world applications, the VOT must periodically interact with an object detector to initiate new tracks, or to validate and/or update the object template with new detector bounding boxes. The quality of bounding boxes produced using a state-of-the-art CNN-based detector can vary, and have an impact on accuracy.  

VOT techniques are mainly classified as either generative or discriminative depending on whether they track by detecting the target or by discriminating the target from the background~\cite{salti2012}. For robust discriminative tracking, adaptive trackers update the target model representation as the object's appearance changes over time.  An adaptive tracker should therefore periodically initiate new tracks (e.g., every second) and update its target representations over time as the object appearance changes. 
\begin{figure}[b]
 \centering
\includegraphics[width=1.0\linewidth]{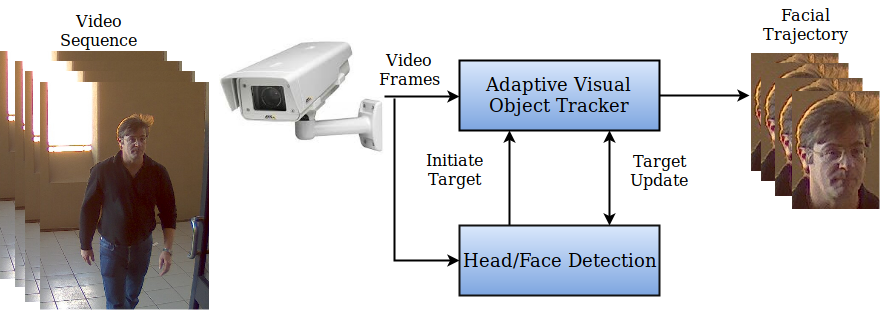}
   \caption{Illustration of the tracker-detector interaction to construct facial trajectories (set of ROIs captured for the same high quality track) in a video surveillance system. Trajectories can be used for further processing, like spatio-temporal person recognition.}
  \label{fig:dt}
\end{figure}

Some techniques have been proposed to combine detection and tracking, and initiate, drop tracks as target objects respectively appear and leave the scene. For example, Fast-DT~\cite{fastdt} relies on detector confidence to drop a track, by efficiently applying the object detector on individual tracker output. Unlike Fast-DT, SORT~\cite{sort},~\cite{MOT_trajectory} focuses mostly on data association and multi-target tracking. Recently methods such as the deep Siamese-FC network~\cite{bertinetto2016fully, SiamRPN} have been proposed to exploit the expressive powers of deep learning in VOT. These SiamFC trackers can effectively learn to represent a target object, but since they do not update the target appearance representations, there is a risk of target drifting due to changes over time in object appearance.  These trackers locate object by finding maximum score location in the output heat map. Hence, when appearance changes abruptly or the object is occluded or partially leaves the search region, the SiamFC tracker temporarily drifts to a location that has a high response map score. Recently, DaSiamRPN~\cite{DaSiam} tracker based on the SiamFC tracking technique further improved by incorporating distracter awareness has produced state of the art results on various tracking benchmarks.

Robust tracking can be achieved by combining a deep detector and tracker. CNN-based object detectors~\cite{8310113} currently provide state-of-the-art accuracy in object detection. However, tracks may drift if the bounding boxes provided by these detector are noisy, due to changes in appearance, background and occlusion. Moreover, given the computational complexity of CNN-based detectors, a key to efficient VOT is the management of detector-tracker interactions. A deep Siamese tracker for real-time video surveillance applications should minimize the number of interactions with the detector for track initiation and update. 

In the literature (e.g.,~\cite{vot2018, otb}), VOT is typically evaluated by initialising the tracker with an initial ground truth target bounding box. These bounding boxes are often tightly bound around the object without much of background noise. In some evaluation methods, bounding boxes are generated with random noise to simulate a practical scenario for tracker initialisation. But these noisy bounding box cannot fully mimic a real-world scenario.

In this paper, the interaction between deep learning models for detection and tracking are analysed with a proposed adaptive tracker. A change detection mechanism is integrated within this Siamese tracker to detect gradual and abrupt changes in a target's appearance in each frame based on features extracted by the deep Siamese network. In response to an abrupt change, the tracker triggers the object detector in order to update an evolving set of templates. Given a gradual change, templates stored in memory are  applied on the search region, and the resulting response maps are integrated to locate a precise target. The proposed  Siamese tracker allows for real-time adaptation of templates, while avoiding target model corruption.

The performance of our adaptive Siamese tracker is compared against baseline Siamese FC trackers, where tracks are initialized and updated with ground truth bounding boxes (ideal object detector) and with the YOLOv3 detector. They are evaluated over several operating conditions on video surveillance like cases from OTB-100~\cite{otb} benchmark where videos contain persons or vehicles.
\begin{figure*}[t!]
 \centering
\includegraphics[width=.7\linewidth]{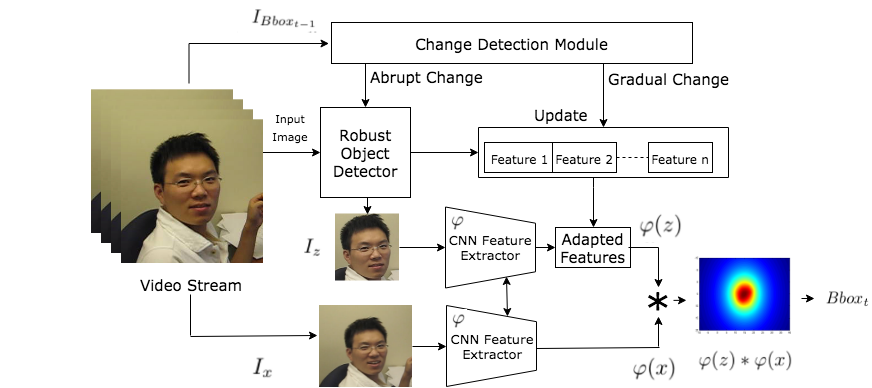}
   \caption{Architecture of the adaptive Siamese tracker that integrates appearance change detection to manage detector-tracker interactions.}
\label{fig:siam}          
\end{figure*}

\section{Tracking Objects in Video Surveillance}

In video surveillance, VOT consists interacts with an object detector. Fig.~\ref{fig:dt} shows an example of the detector-tracker interactions employed  to produce facial trajectories or tracklets. In this case, the face-head detector initiates a new track, and defines a new target representation or template with an initial bounding box. Then, the tracker generates ROIs in subsequent frames. The tracker employs local object detection, learns the object online, and adapts to the changing object appearance and results in the object's location. The detector can also be used locally (on search regions) to validate the tracker's output. Also, in a real-time surveillance application, the detector searches   globally (on the entire frames), and it is often computationally expensive to call the detector every frame.  Objects are tracked by searching locally and can thereby be very  efficient compared to the detector.  

The main challenges of VOT in real-time applications are~\cite{fastdt, 6671560}: (1) tracked objects tend to drift with time due to continuous integration of noise in the target appearance; (2) it is difficult to verify a tracker's state due to lack of reliability in tracker's confidence; (3) the appearance of targets change with time; (4) occlusions are difficult to be detected by the tracker, as there is a risk of learning the occlusion as a part of the target.  Relying on a tracker that is continuously adapting does not guarantee a high-quality trajectory. Trackers that update their template on every frame assuming have a high probability of drifting. Outliers filtering may be  employed in order to detect samples that are notably different from the actual target, and should be removed.



Siamese Fully-Convolutional (SiamFC) tracker~\cite{bertinetto2016fully} uses an AlexNet based Siamese network for feature extraction. The networks takes two input -- target template image $z$ and search image $x$ -- where  $|x| = 2|z|$. The embedding $\varphi$ for $z$ is hence smaller than that of $x$. To localize the object, template features are cross correlated with that of search features to obtain a score map. The location of the maximum value in the score map gives the location of the object in the search region $x$. The tracker had been trained on ILSVRC~\cite{ILSVRC15} dataset with a logistic loss function. During tracking, the correlation map  $f(z, x)$ obtained after cross-correlation of target template embedding $\varphi(z)$ with search image embedding $\varphi(x)$ is defined by:
\begin{equation}
\label{eqn:siam}
f(z, x)=\varphi(z) * \varphi(x)+b 
\end{equation}
In this paper, we focus on deep SiamFC trackers due to their robustness and potential for template adaptation.  Other variants of the Siamese FC tracker have outperformed the baseline SiamFC~\cite{bertinetto2016fully} tracker like SiamRPN~\cite{SiamRPN}, SA-Siam~\cite{SA-Siam}, DaSiamRPN~\cite{DaSiam} and MEMTrack~\cite{MEMtrack}. They have proposed various improvements to the original SiamFC tracker such as attention mechanims, region proposal, etc. to improve overall accuracy and online model adaptation. It is however important in video surveillance to benefit from with the object detector, and adapt to changing appearance.

\section{Adaptive Tracking Using Change Detection}

In our proposed method, a real-time detector is leveraged to initialise and update a deep Siamese tracker with its  object bounding box. In practice, the detector cannot produce a strict bounding box around the object, it is expected to produce bounding boxes that  contain some background context too. Noisy bounding boxes may cause the tracker to quickly drift, and hence it is the tracker template that should be updated when the target appearance begins to change or the tracker starts drifting. As shown in Fig~\ref{fig:siam}, integrating change detection into the Siamese tracker allows to manage detector-tracker interactions.
\begin{figure}[t!]
 \centering
\includegraphics[width=1.0\linewidth]{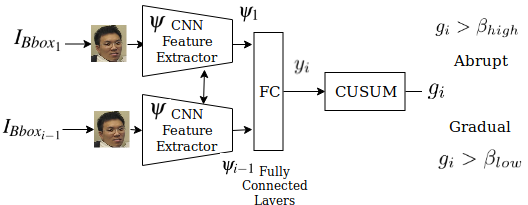}
   \caption{Change detection module with a separate Siamese network from the tracker. The template image is the cropped target image with initial target bounding box and the image to be compared is updated by the current frame cropped by the tracker output bounding box.}
\label{fig:change}          
\end{figure}
\paragraph{Target template similarity measure.}
As a first step to track quality measurement, a Siamese network similar to the SiamFC is employed to predict the similarity measure between current track output and tracker target appearance. The features extracted by the Siamese convolutional layers ($\psi$) are concatenated and a FC layer a shown in Fig.~\ref{fig:change} (Change Detection Module) is trained to predict the similarity score between 0 and 1. The network is trained with positive and negative pairs of object templates from the same video and a random video. Unlike SiamFC tracker, the inputs to the similarity measurement network are cropped along the target bounding box alone without any background context. The feature extraction layer is similar to the one used in SiamFC tracker, with AlexNet features. The extraction of features for the target template occurs just once and is stored in the database. During tracking, it outputs a bounding box location $Bbox_t$ which is used to crop the tracked object image from the current input video stream. This template is then cropped to the CNN input size. Hence the problem of having excess background context as discussed earlier is avoided. In the rest of the paper, this similarity score is referred to as track quality measure.

\paragraph{Change detection with CUSUM.}
Changes in a data stream could be classified as gradual or abrupt depending on the time and magnitude of the change. Hence it has been proposed that in our case the change in appearance of the target would be detected by the template similarity detection module as described above and would be classified as gradual and abrupt. In order to detect change, we propose to use the Adaptive CUmulative SUM algorithm CUSUM~\cite{cusum}. In Eq.~\ref{eqn:mean}, $y_i$ is the tracker similarity measure discussed in the previous section and ~\ref{eqn:statistic}, $g_{i}$ is the test statistic. It is initialised to $0$ at the start of the tracking and when an alarm is detected where $\beta$ is the set threshold. Change is detected when $ g_{i} > \beta $. With a white noise input, it is possible that test statistic will drift away. Hence a small $v$ is subtracted in Equation~\ref{eqn:statistic} to help control the drift. Therefore $v$ and $\beta$ are the design parameters. 
\begin{equation}
\label{eqn:mean}
\hat{\theta}_{i}=\frac{1}{i-i_{0}} \sum_{i=i_{0}+1}^{i} y_{i}
\end{equation}
\vspace{-3mm}
\begin{equation}
\label{eqn:statistic}
g_{i}=\max \left(g_{i-1}-(y_{i}-\hat{\theta}_{i-1})-\nu, 0\right) 
\end{equation}
Using CUSUM for change detection is reliable since it measures changes in similarity score using an evolving statistical model, and comparing each score to a threshold. It allows to measure tracker performance and recognise events. This is different from using a fixed deterministic similarity score because the threshold can depend on object being tracked, and its difficult to set a common threshold.

\paragraph{Model update strategy.}
As opposed to the original SiamFC tracker, the adaptive tracker updates the model dynamically during change detection. A list of tracker models are maintained. These models are selected based on tracker quality scores online during tracking, and are cropped from search regions feature maps. Their search region feature maps are larger than target template feature maps. Hence, during tracking, new models are extracted by cropping feature maps from the search region using location information from tracker output. When a gradual change is detected, i.e indication of a drifting model, the target models or features from the memory are fetched sequentially and matched against the current frame search region to obtain score maps. The final object location is derived from the summation of score maps. This is shown in Algorithm ~\ref{Alg:gradual}.
\vspace{5mm}
\begin{algorithm}[t]
\label{Alg:gradual}
 \KwData{$C$ Memory, search region $ \varphi(x)$}
 \KwResult{$C_{updated} \leftarrow$ final adapted model}
 $integratedScore\leftarrow 0$\;
 $index\leftarrow 0$\;
 \While{ $index <$ length($C$)}{
 
  set current target model as $C[index]$\;
  $f(C[index], x)\leftarrow\varphi(C[index]) * \varphi(x)+b$, track current frame $f_i$\;
  $integratedScore$ = $integratedScore$ + $f(C[index], x)$\;
  $index=index+1$\;
 }
 $BBox_{i} \leftarrow$ get refined bounding box by argmax $integratedScore$\;
 $C_{updated}  \leftarrow$ Crop current template from search region using $BBox_{i}$\;  
 $\varphi_{best}  \leftarrow \varphi(C_{updated})$\; 
 \vspace{2mm}
 \caption{Model update for gradual change.}
\end{algorithm}

Abrupt changes are often caused by occlusion, fast moving objects and in the worst scenario even complete loss of tracking. In these cases it is helpful when a the search region location is reset by an object detector. Since the change is abrupt, it would be necessary for the the target template to be completely re-initialised by the object detector.

\paragraph{Object tracking.} Algorithm ~\ref{Alg:full system} shows the full adaptive Siamese tracking system working with the change detection module.  In the first frame, the tracker is initialised by the object detector. The similarity measurement network is initialised by a cropped target template. During tracking, the change detection module which includes similarity measurement and CUSUM continuously predicts scores and looks for changes in scores. We apply lower and higher thresholds to CUSUM (see Fig.~\ref{fig:change}). The lower threshold detects a gradual change, and higher threshold is to detect an abrupt change. In the case of gradual change (see Algorithm~\ref{Alg:gradual}), the the tracker model is adapted. In case of a detection of abrupt change, indicating drift or drastic change in appearance of the target, the object detector is triggered to detect object and produce bounding boxes to re-initialise the tracking. Algorithm~\ref{Alg:gradual} is called to update memory and adapt target features. Adaptation based on gradual changes is helpful to correct tracking that has just begin drifting or to learn features from latest frames. 

\begin{algorithm}[h!]
\label{Alg:full system}

 \KwData{Image stream $I$, Initial target position $BBox_1$ }
 \KwResult{Estimated target position $BBox_i$}
 
 $i\leftarrow 0$\;
 $scoreMap \leftarrow 0$\;
 $C \leftarrow 0$ \;
 Crop target Image $z$ from  $I_0$ and $BBox_0$\;
 Extract target features $\varphi(z)$\;
 $\varphi_{best} \leftarrow \varphi(z)$\;
 
 \While{ $i < length(I)$}{
  $i\leftarrow i+1$\;
  Extract search region $x_i$ from $I_{i}$ and $BBox_{i-1}$\;
  Extract search embedding $\varphi(x_i)$ from $x_i$\;
  $f(z, x)=\varphi_{best} * \varphi(x)+b$, track current frame $I_i$\;
  $scoreMap = f(z, x)$\;
  estimate track quality measure $y_i$ \;
  \If{$y_i > \alpha$}{
   Extract $\varphi(z_i)$ from  $\varphi(x_i)$ \;
   Add $\varphi(z_i)$ to memory $C_i$\;
   }
  Apply CUSUM on $y_i$ and estimate $g_i$\;
  \If{$g_i > \beta_{low}$}{
   Report gradual change\;
   $C_{updated} \leftarrow $ from Algorithm~\ref{Alg:gradual}\;
   $\varphi_{adapted} \leftarrow C_{updated}$, Update target model\;
   }
   
   \If{$g_i > \beta_{high}$}{
   Report abrupt change detected\;
   $BBox_{detector} \leftarrow$ from Object detector\;
   $\varphi_{detector} \leftarrow$ get embedding from $BBox_{detector}$ and image $I_i$\;
   $\varphi_{temp}\leftarrow \varphi_{detector}$
   reset tracking\;
   $C \leftarrow \varphi_{temp}$ update memory\;
   $\varphi_{best} \leftarrow$ from algorithm 1
   
   }
   \If{$length(C) > Budget$}{
   delete $C[1]$ , remove the oldest but first memory
   }
   $f(z, x)=\varphi_{best} * \varphi(x)+b$\, 
   $scoreMap = f(z, x)$\;
   $BBox_{i} \leftarrow scoreMap$, get final tracker bounding box
 }
 \caption{Algorithm for the overall Adaptive Siamese tracking change detection and model update}
\end{algorithm}

\section{Experimental Methodology}

Proposed and baseline trackers have been evaluated on a subset of videos of OTB-100~\cite{otb} dataset, each one containing person and vehicle being tracked in a video surveillance application.
\begin{table*}[h]
\begin{center}
\scalebox{0.64}
{
\begin{tabular}{|l||r|r|r|r||r|r|r|r||r|r|r|r|}
\hline
                & \multicolumn{12}{|c|}{\bf{Average Overlap (\%)}}     \\ 
\bf{Video}      & \multicolumn{4}{|c||}{\bf{w/o periodic update}} & \multicolumn{4}{|c||}{\bf{w/ periodic update}}  & \multicolumn{4}{|c|}{\bf{w/ periodic update}}   \\ 
                & \multicolumn{4}{|c||}{}  & \multicolumn{4}{|c||}{\bf{(every 30 frames)}}  & \multicolumn{4}{|c|}{\bf{(every 60 frames)}}   \\ 
            &         &      & Adaptive &Adaptive  &     &          &  Adaptive&Adaptive&      &        & Adaptive & Adaptive                \\  
            & SiamFC& DaSiam & Siamese  &DaSiam & SiamFC &DaSiam    &  Siamese& DaSiam &SiamFC &DaSiam  & Siamese  & DaSiam             \\ \hline \hline
Basketball  & 37.95 & 61.27  & 40.27   & 77.49   & 66.19  &71.24    &    41.76& 77.36  &65.30  &78.16    & 69.12  & 73.24    \\ \hline
BlurBody    & 66.72 & 78.28  & 67.01   & 78.30   & 69.64  &78.12    &    67.12& 67.12  &84.59  &74.67    & 82.11  & 78.11    \\ \hline
BlurCar1    & 84.39 & 84.20  & 78.61   & 84.22   & 85.26  &84.15    &    86.73& 84.22  &80.28  &83.50    & 84.35  & 84.15    \\ \hline
BlurCar2    & 84.72 & 83.83  & 84.66   & 83.86   & 85.47  &83.83    &    87.35& 84.03  &81.43  &84.03    & 83.12  & 83.83    \\ \hline
BlurCar3    & 82.89 & 81.73  & 80.52   & 81.79   & 83.40  &85.56    &    88.01& 83.86  &85.73  &83.51    & 82.15  & 85.56    \\ \hline
BlurCar4    & 84.76 & 85.89  & 85.03   & 86.05   & 88.12  &85.99    &    89.15& 85.99  &70.92  &84.48    & 76.72  & 85.99    \\ \hline
Bolt        & 1.17  & 69.68   & 0.64   & 69.64   & 29.95  &77.71    &    80.57& 80.57  &40.68  &76.17     & 44.17 & 77.99      \\ \hline
Bolt2       & 39.77 & 40.15  & 64.14   & 40.47   & 69.25  &69.19    &    84.07& 84.07  &80.43  &65.15    & 87.56  & 77.16     \\ \hline
Car1        & 78.99 & 73.18  & 51.11   & 72.32   & 83.25  &77.27    &    79.69& 79.69  &75.18  &76.35    & 76.11  & 83.88   \\ \hline
Car2        & 88.81 & 82.97  & 88.96   & 84.27   & 88.82  &83.36    &    88.55& 88.55  &78.10  &83.97    & 79.63  & 85.92    \\ \hline
Car24       & 84.63 & 85.18  & 77.56   & 85.21   & 87.18  &85.23    &    88.30& 88.30  &77.66  &79.52    & 77.45  & 83.08    \\ \hline
Car4        & 45.32 & 83.65  & 75.38   & 83.92   & 78.95  &83.72    &    86.64& 86.64  &55.67  &82.91    & 57.31  & 78.21   \\ \hline
CarDark     & 80.96 & 70.27  & 80.55   & 77.03   & 82.96  &77.08    &    82.47& 82.47  &66.53  &75.46    & 68.43  & 77.51      \\ \hline
CarScale    & 65.95 & 72.50  & 75.68   & 75.64   & 70.80  &77.39    &    74.77& 74.77  &66.74  &75.62    & 65.12  & 67.13       \\ \hline
Couple      & 71.49 & 60.49  & 72.25   & 59.74   & 74.46  &59.16    &    76.87& 76.87  &74.82  &56.35    & 77.91  & 71.43      \\ \hline
Crowds     & 63.74  & 63.74  & 7.64    & 67.06   &64.42   &66.63    &    76.64& 76.64  &71.67  &67.36    & 73.83  & 70.45       \\ \hline
David3     & 50.68  & 70.45  & 70.70   & 70.70   & 72.36  &70.45    &    66.04& 66.04  &58.00  &69.14    & 57.25  & 56.30      \\ \hline
Diving     & 10.80  & 53.64  & 23.29   & 57.09   & 23.35  &56.89    &    35.01& 35.01  &65.30  &56.13    & 67.15  & 71.75   \\ \hline
Girl2      & 64.26  & 64.48  & 61.32   & 61.28   & 69.95  &64.46    &    75.24& 75.24  &67.66  &61.76    & 66.87  & 76.48    \\ \hline
Human2     & 75.25  & 76.58  & 70.32   & 76.68   & 79.10  &76.48    &    77.44& 77.44  &69.01  &74.74    & 74.78  & 76.08    \\ \hline
Human3     &  1.41  &  69.60   & 20.60 & 71.79   & 59.44  &76.08    &    62.46& 62.46  &69.59  &75.81    & 77.66  & 76.63      \\ \hline
Human4     & 33.45  & 36.63  & 32.58   & 37.48   & 56.79  &36.63    &    61.70& 61.70  &76.45  &41.44    & 62.37  & 56.46    \\ \hline
Human5     & 51.33  & 78.36  & 75.34   & 79.17   & 77.88  &78.29    &    82.57& 82.57  &2.41   &78.78    & 73.47  & 80.68      \\ \hline
Human6     & 74.93  & 73.77  & 74.02   & 75.81   & 77.65  &73.85    &    75.56& 75.56  &74.31  &71.71    & 66.85  & 73.48      \\ \hline
Human7     & 74.77  & 74.43  & 75.11   & 75.94  & 79.60  &74.97    &    81.33&  81.33  &64.09   &33.77    & 77.74 & 69.08      \\ \hline
Human8     & 6.00   & 70.12   & 76.75  & 72.18   & 41.93  &69.84    &    76.75& 76.75  &78.38  &67.43    & 74.92  & 69.84   \\\hline
Human9     & 69.94  & 71.25  & 69.01   & 72.48   & 73.77  &71.53    &    81.95& 81.95  &70.61  &82.14    & 73.51  & 71.53    \\ \hline
Jogging-1  & 69.78  & 58.75  & 68.89   & 74.34   & 71.22  &58.96    &    78.12& 78.12  &73.50  &71.26    & 48.86  & 68.96     \\ \hline
Jump       & 23.55  & 49.05  & 13.08   & 49.81   & 25.16  &49.03    &    40.64& 40.64  &31.38  &51.60    & 76.32  & 49.03    \\ \hline
RedTeam    & 68.58  & 74.71  & 65.96   & 74.17   & 74.93  &74.45    &    74.15& 74.15  &69.62  &40.06    & 82.44  & 79.45    \\ \hline
Singer1    & 77.13  & 71.97  & 77.52   & 72.59   & 79.30  &71.87    &    83.48& 83.48  &79.01  &77.98    & 65.94  & 77.87    \\ \hline
Singer2    &  3.59  &  30.20   & 30.45 & 59.32   & 63.21  &38.16    &    79.32& 79.32  &65.94  &37.87    & 79.71  & 55.16     \\ \hline
Skater     & 64.38  & 70.69  & 64.05   & 71.55   & 60.54  &70.69    &    62.89& 62.89  &76.22  &60.26    & 61.85  & 70.69       \\ \hline
Skater2    & 59.21  & 70.14  & 60.08   & 72.28   & 61.49  &70.11    &    66.10& 66.10  &67.19  &68.47    & 64.18  & 70.11      \\ \hline
Skating1   & 23.72  & 68.71  & 36.87   & 64.09   & 39.41  &68.38    &    42.18& 42.18  &80.65  &65.38    & 39.15  & 68.38   \\ \hline
Subway     & 17.47  & 35.14  & 18.49   & 37.64   & 66.95  &63.91    &    68.77& 68.77  &48.58  &61.81    & 66.71  & 63.91      \\ \hline
Suv        & 64.55  & 66.38  & 64.65   & 66.71   & 77.51  &46.68     &   83.16& 83.16  &79.64  &74.78    & 65.92  & 76.68     \\ \hline
Walking    & 75.49  & 69.95  & 71.68   & 71.12   & 73.92  &69.94    &    74.93& 74.93  &28.19  &71.93    & 75.69  & 73.94      \\ \hline
Walking2   & 49.36  & 26.90  & 33.65   & 29.98   & 79.20  &77.46    &    77.49& 77.49  &65.95  &77.57    & 71.66  & 77.46   \\ \hline
Woman      & 13.10  & 56.93  & 43.02   & 59.83   & 51.80  &64.18    &    70.74& 70.74  &67.11  &63.45    & 70.19  & 64.18 \\\hline\hline
\bf{Average} & 54.32\ &66.59   & 57.93   & 69.02   & 67.13  &70.75     &   73.96& 75.94  &68.92  &69.06   & 71.18  & 73.73     \\ \hline
\end{tabular}
}
\end{center}
\caption{Average overlap of the original SiamFC, DaSiamRPN(DaSiam) and proposed Adaptive Siamese trackers on OTB-100 subset videos. Results are shown with and without periodic update (every 30 and 60 frames) of the tracker template using ground truth. The templates have been initialised on the first frame using the ground truth bounding box. In the periodic cases, ground truth bounding boxes are also employed for the object being tracked to update the template.}
\label{tab:gt_init}
\end{table*}
For the tracker,  original training and implementation of the SiamFC~\cite{bertinetto2016fully} and DaSiamRPN~\cite{DaSiam} tracker was reproduced. The track quality measurement network was trained using ILSVRC2015~\cite{ILSVRC15} dataset similar to SiamFC. This was trained as a similarity measurement network to compare tracked object image with that of a template image that initialised the tracker. Each batch for training would consist 8 pairs of positive and negative samples. Positive samples were generated  similar to SiamFC and negative samples were either other images of other objects or background images around the target. The network was trained using a logistic loss function. YOLOv3~\cite{yolov3} object detector trained on COCO dataset was used to initialize the tracker. 

Two sets of experiments were performed on the Siamese trackers, one with initialization from ground truth bounding box and the other initialised by object detector's bounding box. In a typical surveillance system as shown in Fig.~\ref{fig:dt}, the object detector would be called every $N$ frames to discover possible appearances of new objects and begin tracking. At the same time, the detector output would contain objects from corresponding videos or tracks. This would be used to update the tracks. We call this a periodic update. In order to associate the detector bounding boxes with existing track bounding boxes, some of the common strategies include using Intersection Over Union (IOU) to update, or using a motion model, etc. Hence in our implementation, we propose to use IOU and a constant velocity motion model with a Kalman Filter to update a tracker with detection bounding boxes and also to search for an object with a detector similar to ~\cite{sort}. 

\begin{table*}[t]
\begin{center}
\scalebox{0.65}
{
\begin{tabular}{|l||c||r|r|r|r||r|r|r|r||r|r|r|r|}
\hline
                & \multicolumn{13}{|c|}{\bf{Average Overlap (\%)}}     \\ 
\bf{Video} &{\bf{Detector}}      & \multicolumn{4}{|c||}{\bf{w/o periodic update}} & \multicolumn{4}{|c||}{\bf{w/ periodic update}}  & \multicolumn{4}{|c|}{\bf{w/ periodic update}}   \\ 
               &{\bf{Performance}} & \multicolumn{4}{|c||}{}  & \multicolumn{4}{|c||}{\bf{(every 30 frames)}}  & \multicolumn{4}{|c|}{\bf{(every 60 frames)}}   \\ 
                       &    &     &      & Adaptive &Adaptive  &     &          &  Adaptive&Adaptive&      &                                    & Adaptive  & Adaptive                \\  
            &           & SiamFC    & DaSiam    &   Siamese &  DaSiam   &  SiamFC   &    DaSiam &   Siamese &  DaSiam   &   SiamFC  &   DaSiam  & Siamese   & DaSiam             \\ \hline \hline
Basketball 	&	30.08	&	1.47	&	21.11	&	8.98	&	26.46	&	2.27	&	26.49	&	3.76	&	32.46	&	1.34	&	14.63	&	2.12	&	16.53	 \\	\hline	
BlurBody   	&	76.43	&	74.47	&	72.18	&	77.95	&	77.41	&	72.16	&	76.52	&	75.30	&	79.41	&	74.29	&	77.90	&	79.62	&	76.15	 \\	\hline	
BlurCar1   	&	85.94	&	81.01	&	82.41	&	78.59	&	82.98	&	77.76	&	83.46	&	79.52	&	85.59	&	78.47	&	79.38	&	74.83	&	84.15	 \\	\hline	
BlurCar2   	&	34.69	&	36.12	&	41.55	&	82.31	&	56.86	&	70.20	&	63.86	&	40.63	&	64.09	&	78.76	&	59.34	&	59.25	&	62.78	\\	\hline	
BlurCar3   	&	21.79	&	26.15	&	31.76	&	76.12	&	37.43	&	71.96	&	38.43	&	25.69	&	41.26	&	70.07	&	35.82	&	34.74	&	37.14	\\	\hline	
BlurCar4   	&	89.41	&	82.21	&	81.53	&	81.70	&	82.55	&	81.42	&	82.78	&	82.00	&	84.00	&	80.47	&	82.00	&	83.01	&	83.29	 \\	\hline	
Bolt       	&	60.78	&	1.04	&	18.33	&	1.04	&	22.78	&	8.90	&	24.12	&	18.09	&	27.14	&	1.04	&	16.06	&	12.95	&	12.76	 \\	\hline	
Bolt2      	&	62.18	&	37.52	&	37.21	&	56.04	&	39.95	&	48.40	&	39.95	&	37.11	&	42.02	&	39.11	&	27.85	&	42.86	&	38.13	\\	\hline	
Car1       	&	87.25	&	71.25	&	72.44	&	73.78	&	73.21	&	72.33	&	73.21	&	77.72	&	74.41	&	72.75	&	65.58	&	80.08	&	71.17	\\	\hline	
Car2       	&	70.94	&	70.49	&	74.92	&	81.49	&	78.34	&	79.99	&	78.34	&	72.38	&	78.26	&	77.75	&	72.26	&	80.08	&	75.21	\\	\hline	
Car24      	&	75.86	&	48.98	&	53.83	&	84.57	&	58.47	&	86.79	&	67.47	&	42.58	&	68.91	&	84.81	&	42.03	&	79.18	&	63.4	\\	\hline	
Car4       	&	92.63	&	63.48	&	67.65	&	83.00	&	71.26	&	81.73	&	71.26	&	73.78	&	73.78	&	80.44	&	73.78	&	76.18	&	68.21	\\	\hline	
CarDark    	&	28.91	&	17.11	&	32.12	&	1.72	&	22.46	&	29.18	&	22.46	&	26.72	&	23.60	&	27.35	&	17.97	&	11.62	&	12.38	\\	\hline	
CarScale   	&	92.10	&	63.54	&	61.66	&	69.14	&	66.64	&	68.45	&	69.64	&	62.70	&	74.63	&	69.26	&	66.67	&	69.85	&	67.78	\\	\hline	
Couple     	&	47.69	&	53.20	&	54.57	&	36.20	&	59.74	&	54.69	&	59.74	&	57.35	&	61.58	&	54.54	&	52.75	&	57.66	&	58.17	\\	\hline	
Crowds     	&	33.12	&	1.33	&	12.53	&	1.33	&	15.83	&	5.48	&	46.83	&	54.64	&	53.39	&	1.33	&	23.29	&	54.51	&	51.23	\\	\hline	
David3     	&	2.78	&	0.45	&	7.35	&	0.62	&	17.67	&	2.28	&	17.67	&	6.16	&	19.67	&	0.63	&	19.74	&	4.02	&	23.31		 \\	\hline	
Diving     	&	72.05	&	8.44	&	14.57	&	15.05	&	18.72	&	15.41	&	28.72	&	19.25	&	15.80	&	12.91	&	15.60	&	13.13	&	12.27		 \\	\hline	
Girl2      	&	41.11	&	0.07	&	16.64	&	31.86	&	44.16	&	21.07	&	44.16	&	20.79	&	45.35	&	23.49	&	27.68	&	20.13	&	41.76		 \\	\hline	
Human2     	&	84.20	&	72.96	&	71.42	&	22.77	&	74.63	&	45.06	&	74.63	&	47.22	&	75.17	&	43.11	&	33.40	&	42.57	&	71.44		 \\	\hline	
Human3     	&	12.46	&	0.09	&	4.12	&	17.99	&	16.75	&	15.73	&	22.75	&	41.45	&	23.47	&	0.08	&	37.19	&	38.59	&	22.72		 \\	\hline	
Human4     	&	33.13	&	1.09	&	5.25	&	11.58	&	11.37	&	26.85	&	11.37	&	31.27	&	14.29	&	8.52	&	27.84	&	4.62	&	12.29		 \\	\hline	
Human5     	&	56.42	&	0.14	&	3.66	&	0.14	&	7.62	&	3.69	&	7.62	&	58.17	&	16.00	&	0.14	&	41.19	&	55.62	&	15.78		 \\	\hline	
Human6     	&	71.84	&	1.49	&	12.59	&	43.67	&	18.31	&	57.94	&	18.31	&	51.82	&	27.72	&	39.72	&	49.57	&	17.83	&	22.16		 \\	\hline	
Human7     	&	88.21	&	76.56	&	72.17	&	81.10	&	73.72	&	73.97	&	73.72	&	21.96	&	74.62	&	75.25	&	76.64	&	78.02	&	71.41		 \\	\hline	
Human8     	&	89.48	&	5.51	&	7.46	&	46.56	&	11.39	&	20.91	&	44.39	&	77.64	&	45.81	&	24.52	&	77.09	&	75.49	&	44.79		 \\	\hline	
Human9     	&	48.13	&	40.38	&	32.75	&	79.52	&	45.35	&	45.39	&	69.35	&	43.22	&	71.52	&	73.99	&	67.63	&	45.73	&	64.11		 \\	\hline	
Jogging-1  	&	11.85	&	12.86	&	22.93	&	57.98	&	66.11	&	21.77	&	62.17	&	42.90	&	62.95	&	47.87	&	39.77	&	41.52	&	59.37		 \\	\hline	
Jump       	&	20.07	&	2.62	&	1.44	&	2.35	&	15.54	&	15.57	&	14.62	&	32.78	&	15.92	&	2.35	&	25.87	&	28.66	&	14.26		 \\	\hline	
RedTeam    	&	73.82	&	66.61	&	71.31	&	43.87	&	71.76	&	41.10	&	74.79	&	28.58	&	75.38	&	40.67	&	29.17	&	52.46	&	73.32		 \\	\hline	
Singer1    	&	65.39	&	2.06	&	17.72	&	48.49	&	25.16	&	52.03	&	42.16	&	66.33	&	43.71	&	0.35	&	67.11	&	1.08	&	33.23		 \\	\hline	
Singer2    	&	29.20	&	1.59	&	17.31	&	33.43	&	28.11	&	11.92	&	37.11	&	21.23	&	38.98	&	10.71	&	32.18	&	29.90	&	33.47		 \\	\hline	
Skater     	&	44.39	&	49.70	&	61.74	&	60.21	&	63.83	&	61.12	&	67.83	&	49.50	&	68.35	&	62.84	&	18.14	&	51.16	&	55.22		 \\	\hline	
Skater2    	&	59.66	&	58.66	&	61.38	&	65.29	&	65.28	&	25.05	&	44.28	&	50.30	&	46.30	&	61.06	&	50.30	&	54.59	&	37.35		 \\	\hline	
Skating1   	&	6.12	&	2.20	&	5.73	&	44.59	&	7.84	&	49.76	&	39.84	&	58.21	&	41.71	&	18.45	&	54.82	&	21.43	&	40.28		 \\	\hline	
Subway     	&	36.45	&	8.90	&	11.25	&	10.29	&	18.27	&	14.99	&	18.27	&	5.05	&	19.06	&	8.90	&	10.05	&	10.28	&	14.71		 \\	\hline	
Suv        	&	71.85	&	59.92	&	61.36	&	38.78	&	64.86	&	60.46	&	58.11	&	59.43	&	57.86	&	39.41	&	53.03	&	58.75	&	46.38		 \\	\hline	
Walking    	&	24.17	&	0.24	&	3.11	&	3.44	&	8.12	&	1.22	&	28.17	&	10.24	&	30.25	&	0.60	&	16.24	&	0.24	&	31.72		 \\	\hline	
Walking2   	&	39.26	&	1.39	&	11.9	&	22.55	&	19.94	&	21.83	&	13.56	&	12.44	&	18.45	&	19.23	&	7.06	&	0.70	&	16.66		 \\	\hline	
Woman      	&	55.84	&	54.58	&	50.33	&	62.39	&	56.53	&	47.32	&	53.53	&	56.00	&	58.47	&	62.43	&	55.80	&	57.98	&	62.36		 \\	\hline	\hline

\bf{Average}&-	&	31.45	&	36.53 	&	43.46	&	43.09	&	41.61	&	47.30	&	43.55	&	49.28	&	39.23	&	43.51	&	42.58	&	44.97 
     \\ \hline
\end{tabular}
}
\end{center}
\caption{Average overlap of the original SiamFC, DaSiamRPN (DaSiam) and proposed Adaptive Siamese trackers on OTB-100 subset videos.  Results are shown with and without periodic update of the tracker template using object detector.  The templates have been initialised on the first frame using the YOLOv3 object detector as in a real world video surveillance scenario (Instead of ground truth unlike in OTB evaluations and in Tab 1).}
\label{tab:detector_init}
\end{table*}

\begin{figure*}[h]
\centering
\begin{tabular}{ll}
\includegraphics[scale=0.3]{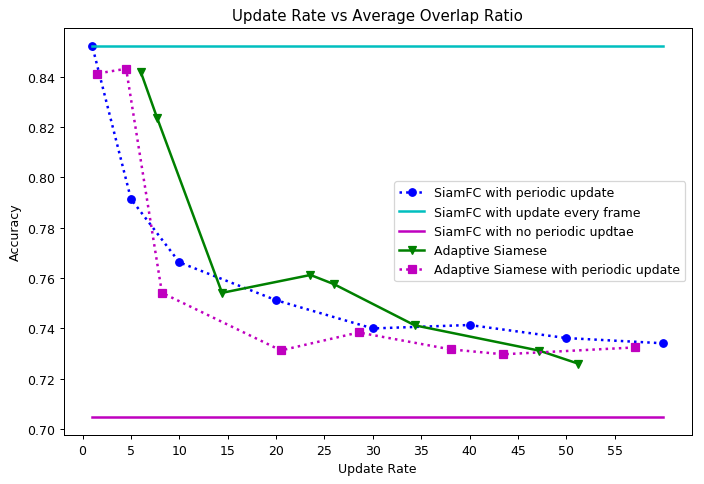}
&
\includegraphics[scale=0.3]{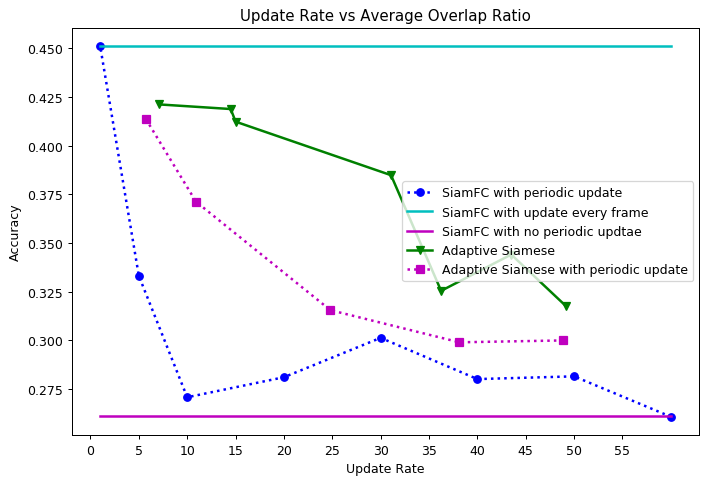}
\end{tabular}
\caption{\textbf{Left}: Analysis of accuracy versus update rate for different up-date strategies for tracker template with the car2 video from OTB-100 dataset, where detector initialization has a good IOU w.r.t. theground truth(70\%) and \textbf{Right}: same as in left but with Skating2 video from OTB-100 dataset}
\label{Fig:analysis}
\end{figure*}

The measures used for evaluation performance are: (1) the OTB benchmark measure (percentage of frames with IOU $>$ 0.5), and (2) the average overlap $\overline{\phi}$ between ground truth and tracked bounding boxes over all $N$ frames in a video sequence: 
\begin{equation}
\label{eqn:IOU}
\overline{\phi} = \frac{1}{N} \sum_{t} \phi_{t} = \frac{1}{N} \sum_{t} \frac{R_{t}^{G} \cap R_{t}^{T}}{R_{t}^{G} \cup R_{t}^{T}}
\end{equation}
where $R_{t}^{G}$ and $R_{t}^{T}$ are the ground truth and tracked bounding box regions, and $t = 1, 2, ..., N$ frames in a video.

\section{Results and Discussion}

Two sets of experiments were performed -- in the first case the  trackers interact with an ideal detector (ground truth bounding boxes of the object), and in the second case with a real-world YOLOv3~\cite{yolov3} object detector. The corresponding results are shown Tabs.~\ref{tab:gt_init} and~\ref{tab:detector_init}. Column 2 of Tab.~\ref{tab:detector_init} shows the IOU of the object detector's bounding box and the ground truth bounding box of the object in the first frame. From the tables it can be observed that videos that had a poor detector IOU (with respect to ground truth) during initialization have an overall lower performance compared to that of the videos from the ground truth initialization. The VOT performance of deep Siamese trackers declines significantly when using the real-world object detector.  However, adaptive Siamese tracking can improve tracking performance by enabling effective detector-tracker interactions especially in videos where ground truth error is high. The average speed of our Adaptive DaSiamRPN is $71 \pm 15$ fps and that of Adaptive Siamese is $55 \pm 12$ fps exhibiting real time performance.

The horizontal axis of Fig.~{4} represents the average update rate for a video obtained by changing the threshold of change detection module to be more or less sensitive to changes. This alters the average rate at which the tracker is updated over an entire video. This has helped us to compare different update strategies over accuracy and complexity in terms of the average number times a detector is called over the entire length of a video.
In order to analyse the interaction between noisy detections and tracker, two special cases have been selected from the OTB-100 sequences. In one case i.e car2 sequence, the detector bounding box IOU with that of ground truth bounding box of object is good i.e  70\%. Hence in this case, it can be observed that the Adaptive Tracking or AdpSiam (ours) tracker, SiamFC with a periodic update are almost the same with a very marginal advantage from AdpSiam. This is due to low error in the initialization and also, the car2 sequence is an easy case with a steady movement of the object.

The second case (skating2) is a more complex video from OTB-100 dataset, where the detector bounding box has a 21\% IOU with the ground truth bounding box. Note that this is a person tracking video and hence the aspect ratio is greater than one. The target template would hence contain greater amount of noise as compared to the car2 sequence. This has caused the SiamFC with periodic update to perform poorly compared to AdpSiam with periodic update and AdpSiam. Also, the video in itself is complex with a few instances of occlusion background clutter. 

\begin{figure*}[h!]
 \centering
\includegraphics[width=0.74\linewidth]{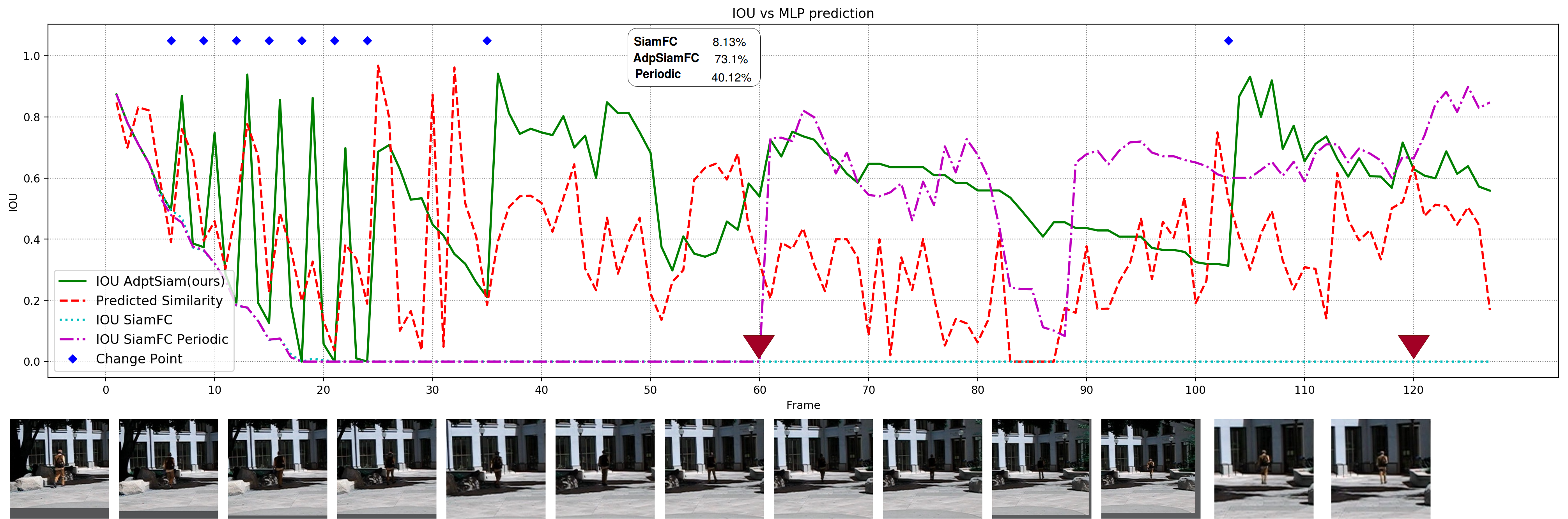}
   \caption{Frame-by-frame analysis of tracker performance (IOU over time) of the trackers on the Human8 video of OTB-100 dataset for each model update strategy.}
\label{fig:hum}          
\end{figure*}
A frame-by-frame analysis of tracking performance is shown in Fig.~\ref{fig:hum}. The horizontal axis shows the frame count, and the vertical axis shows the corresponding tracker accuracy in terms of IOU with ground truth. The performance of the original SiamFC is poor in this particular video (Human8) of OTB-100 dataset. It can be seen that the tracker drifts completely at frame number eighteen. The sequence was also evaluated in the case of SiamFC tracker with a periodic update of sixty frames. This is indicated by the inverted triangle on the horizontal axis. The performance of periodic update is lower than that of the adaptive update because a fixed period of 60 fails to stop the tracker drift at the eighteenth frame. Hence adaptive detection and tracking optimise the update rate for a given video. At the same time, Note that the number of updates for the adaptive tracker is higher in this case due to the complexity of the video. With the adaptive tracker, it is possible to dynamically adjust the update rate without the need of a threshold for number of updates. The same is true for a video with fewer appearance changes. In this case, adaptive tracker reduces the number of updates required while tracking based on the periodic update to the tracker would update the tracker periodically irrespective of the necessity of update.

\section{Conclusion}

In this paper, the interaction between deep learning models for detection and tracking are analysed. An adaptive Siamese tracker has been introduced that leverages a change detection mechanism to manage its interactions with a detector in real-world video surveillance applications. However, in practice object detectors are noisy, and therefore the tracks initialized with the corresponding bounding boxes tend to drift rapidly. Given the detection of an abrupt appearance change, the proposed tracker relies on the object detector to re-initialize the track template, while for gradual change detection, the detector is used to update an evolving set of templates. 
Results on the videos from the OTB-100 dataset highlight the importance of detection in long-term VOT -- the tracking performance can decline considerably even with state-of-the-art deep YOLOv3 detector. In all cases, there is a clear benefit in updating templates on a periodic basis. The proposed adaptive Siamese tracker always outperforms the original Siamese FC and DaSiamRPN trackers especially in cases where the tracker initialization is associated with high object detection error. Using change detection allows adapting the update rate to the challenges encountered during tracking, as opposed to using only a fixed periodic update rate. This enables detector-tracker interactions that do not rely on heuristics to update the tracker templates. In future research, track quality measurements will be explored to  further improved performance by training with realistic noisy samples similar to those from an object detector's output to improve immunity to noisy detector initialization.
%
%


\FloatBarrier
{\small
\bibliographystyle{ieee}
\bibliography{egbib}
}


\end{document}